\documentclass[11pt]{article}

\usepackage[preprint]{acl}

\usepackage{times}
\usepackage{latexsym}
\usepackage{booktabs} 
\usepackage{multirow}
\usepackage[T1]{fontenc}

\usepackage[utf8]{inputenc}

\usepackage{microtype}

\usepackage{inconsolata}

\usepackage{graphicx}

\title{Scaling Multilingual Safety and Reasoning Evaluation through Reasoning-Guided Translation}

\author{
  Zahraa Al Sahili\textsuperscript{1}\thanks{\ \ Corresponding author.} \quad
  Ioannis Patras\textsuperscript{1} \quad
  Matthew Purver\textsuperscript{1,2} \\
  \textsuperscript{1} School of Electronic Engineering and Computer Science, \\ Queen Mary University of London, London, UK \\
  \textsuperscript{2} Department of Knowledge Technologies, \\ Jožef Stefan Institute, Ljubljana, Slovenia \\
  \texttt{\{z.alsahili, i.patras, m.purver\}@qmul.ac.uk}
}

\usepackage{xcolor}

\usepackage{fvextra}
\usepackage{xurl}

\newcommand{\nUntranslated}{274}
\newcommand{\nLabelJudg}{14{,}556} 
\newcommand{\nExcludedTriples}{370}

\begin{document}
\maketitle
\begin{abstract}
Large language models (LLMs) are increasingly deployed across languages, yet their safety and reasoning capabilities remain predominantly evaluated in English. Translating existing benchmarks offers a scalable approach to multilingual evaluation, but preserving task-specific meaning across languages presents significant challenges, particularly for harmful requests, mathematical reasoning, and conversational interactions. We introduce a reasoning-guided translation framework for constructing multilingual evaluations that preserve the structure and intended meaning of diverse benchmark tasks. Our approach separates translation strategy selection from large-scale generation, combining reasoning-guided configuration selection with task-aware validation, targeted repair, and independent quality screening. Using this framework, we develop a source-aligned evaluation collection spanning seven safety, conversational, and mathematical reasoning resources across 11 languages. The resulting collection achieves 99.94\% coverage of intended item--language pairs in the two subsets requiring repair. Evaluations of open-source LLMs reveal distinct language-dependent vulnerabilities, including inconsistent refusal behaviour and substantial reasoning degradation in lower-resource languages. These findings highlight the importance of task-sensitive benchmark construction and systematic quality assessment for understanding multilingual model behaviour beyond English.
\end{abstract}

\section{Introduction}
\label{sec:introduction}

Large language models (LLMs) are increasingly used across
linguistic communities, yet their safety and reasoning
capabilities remain predominantly evaluated in English
\citep{yong-etal-2025-state}. This disparity raises
concerns about the reliability of multilingual deployment,
as model behaviour can vary substantially across
languages. Previous studies have shown that language
choice can influence responses to harmful requests
\citep{deng2024multilingual} and that reasoning performance
often differs across linguistic settings
\citep{shi2023multilingual,son-etal-2025-linguistic}.
However, interpreting these differences requires
comparable evaluation resources. Without equivalent
test instances, observed performance gaps may reflect
differences in benchmark construction rather than
the underlying capabilities of the models.

Translating established English benchmarks provides
a practical and scalable means of extending evaluation
to other languages. However, linguistic fluency alone
does not guarantee that translated instances preserve
the properties being evaluated. Safety benchmarks
require the original intent and severity of harmful
requests to remain intact, while maintaining distinctions
between benign and unsafe content. Mathematical
reasoning tasks demand precise preservation of numerical
relationships and logical constraints, whereas
conversational evaluations depend on consistent dialogue
structure and message ordering. These challenges
are further complicated by the safety alignment of
translation models, which may refuse to translate
harmful-looking requests or alter their intended meaning.
Such behaviour represents a potential source of
evaluation bias, particularly when unsuccessful
translations are silently excluded. Consequently,
reliable multilingual benchmark construction requires
more than translation accuracy: it demands task-aware
quality control, transparent handling of failures,
and systematic assessment of the resulting resources.

\begin{figure*}[t]
    \centering
    \includegraphics[width=0.7\textwidth]{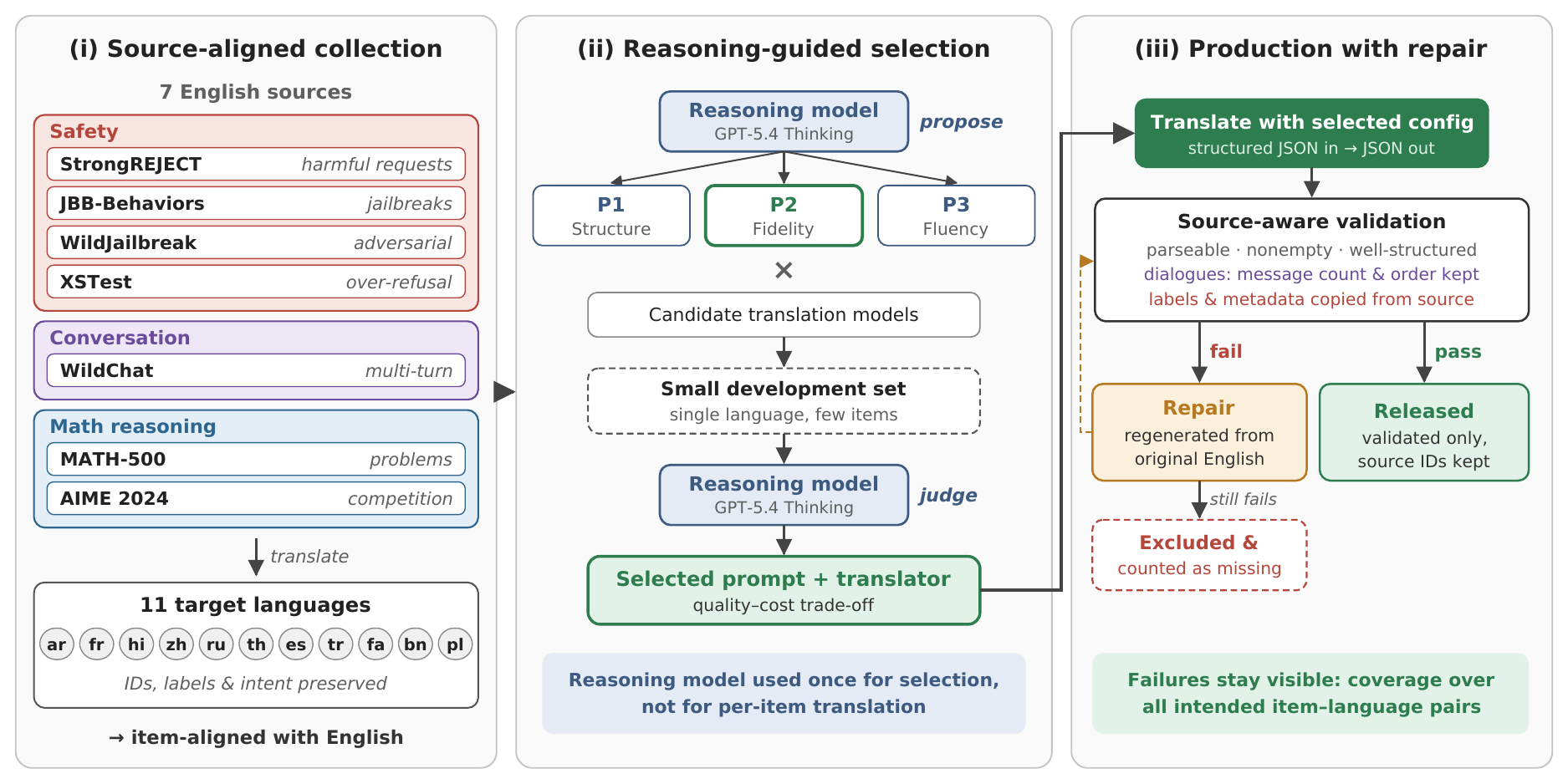}
    \caption{Overview of the reasoning-guided multilingual
    benchmark construction framework. (i) Seven English
    resources are adapted into 11 target languages while
    retaining source identifiers and task-specific
    annotations. (ii) A reasoning model proposes and
    evaluates translation configurations before
    large-scale generation. (iii) Source-aware validation
    and targeted repair address structural failures,
    while unresolved cases are explicitly recorded.}
    \label{fig:overview}
\end{figure*}

To address these challenges, we introduce a
\textbf{reasoning-guided translation framework}
for scalable multilingual benchmark construction.
Our approach separates translation configuration
design from large-scale generation, leveraging a
reasoning model to identify suitable translation
strategies while delegating production to a more
computationally efficient model. Unlike a uniform
translation pipeline, the proposed framework accounts
for the distinct requirements of safety, conversational,
and mathematical reasoning tasks. It integrates
source-aware validation, targeted repair, and
translator-independent quality screening to identify
potential translation failures and maintain traceability
to the original evaluation instances. Figure
\ref{fig:overview} illustrates the overall workflow.

Using this framework, we construct a source-aligned
multilingual evaluation collection spanning seven
English-language resources and 11 target languages.
The collection brings together complementary
evaluation settings, including harmful requests
(StrongREJECT and JBB-Behaviors), adversarial and
benign safety prompts (WildJailbreak), over-refusal
(XSTest), naturalistic conversations (WildChat),
and mathematical reasoning (MATH-500 and AIME~2024).
By retaining source identifiers and task-specific
annotations, the resulting collection enables
item-level comparisons across languages while
preserving the distinct evaluation purposes of
the original benchmarks.

Our evaluation demonstrates the practical effectiveness
of the framework in constructing structurally valid
multilingual resources, achieving 99.94\% coverage
of intended item--language pairs in the two subsets
requiring repair. Downstream evaluations further
reveal substantial differences in how LLMs respond
to multilingual safety and reasoning tasks, including
model-specific refusal patterns and performance
degradation in lower-resource languages. These
findings underscore the importance of evaluating
multilingual capabilities using aligned, systematically
validated resources rather than relying exclusively
on English-centric benchmarks or aggregate performance
measures.

Our main contributions are as follows:

\begin{itemize}
    \item We construct a \textbf{source-aligned multilingual
    evaluation collection} spanning seven safety,
    conversational, and mathematical reasoning resources
    across 11 languages, enabling consistent item-level
    comparisons with their English counterparts.

    \item We introduce a \textbf{reasoning-guided translation
    framework} that separates translation strategy
    selection from large-scale generation, supported
    by a systematic configuration study examining
    translation validity, efficiency, and cost.

    \item We develop an \textbf{auditable validation and
    repair workflow} incorporating source-specific
    structural checks, targeted regeneration, and
    translator-independent screening to identify
    potential fidelity issues and account for
    unsuccessful translations.

    \item We provide a \textbf{cross-lingual evaluation
    of LLM safety and reasoning}, revealing distinct
    model-dependent vulnerabilities and highlighting
    the importance of multilingual evaluation for
    understanding model behaviour beyond English.
\end{itemize}

\section{Related Work}
\label{sec:related-work}

\paragraph{Multilingual safety evaluation.}
Existing safety research disproportionately focuses on English, leaving important gaps in non-English evaluation \citep{yong-etal-2025-state}. XSafety measures harmful-response behavior across ten languages \citep{wang-etal-2024-languages}, while \citet{deng2024multilingual} show that multilingual prompting can expose vulnerabilities in safety alignment. At the benchmark level, JailbreakBench establishes standardized harmful behaviors and evaluation protocols for jailbreak robustness \citep{chao2024jailbreakbench}. These studies motivate language-aligned evaluation, but do not jointly cover our safety, over-refusal, and conversational sources.

\paragraph{Multilingual mathematical reasoning.}
MGSM introduced manually translated grade-school mathematics problems for cross-lingual reasoning evaluation \citep{shi2023multilingual}. More recent benchmarks target competition-level mathematics: MCLM includes translated AIME 2024 and MATH problems across a broad range of languages \citep{son-etal-2025-linguistic}, and MMATH combines model-generated translations, cross-model verification, and human revision \citep{luo-etal-2025-mmath}. We therefore do not claim the first translations of these tasks; instead, we integrate them with complementary safety sources in one workflow.

\paragraph{Benchmark translation and automated selection.}
Global MMLU demonstrates that translated evaluations can retain linguistic and cultural biases \citep{singh-etal-2025-global}; MuBench further emphasizes cross-language alignment and translation-quality assessment \citep{han-etal-2026-mubench}. Automatic Prompt Engineer and ProTeGi establish LLM-based instruction proposal and optimization \citep{zhou2023ape,pryzant-etal-2023-automatic}, while M-MAD explores structured LLM judging for translation quality \citep{feng-etal-2025-mad}. We build on these directions by using a reasoning model to select a translation configuration on a small development set and a lighter model for production, while distinguishing format checks from semantic validation.

\section{Benchmark Construction}
\label{sec:benchmark}

We construct a multilingual evaluation collection from
seven English-language resources encompassing complementary
aspects of model safety, conversational behaviour, and
mathematical reasoning (Table~\ref{tab:sources}).
StrongREJECT \citep{souly2024strongreject} and
JBB-Behaviors \citep{chao2024jailbreakbench} provide
harmful requests, while WildJailbreak
\citep{jiang2024wildteaming} includes adversarial and
benign safety-related prompts. XSTest
\citep{rottger-etal-2024-xstest} evaluates over-refusal,
and WildChat \citep{zhao2024wildchat} contributes
naturally occurring user--assistant conversations.
For mathematical reasoning, we incorporate MATH-500
and AIME~2024
\citep{lightman2023verify,son-etal-2025-linguistic}.
These resources capture distinct evaluation objectives,
allowing multilingual comparisons across several
dimensions of model behaviour without conflating
them into a single performance measure.

The collection covers 11 target languages: Arabic (ar),
French (fr), Hindi (hi), Chinese (zh; Simplified script),
Russian (ru), Thai (th), Spanish (es), Turkish (tr),
Persian (fa), Bengali (bn), and Polish (pl).
To maintain alignment with the English sources,
we translate natural-language fields while retaining
record identifiers, metadata, mathematical notation,
reference answers, and code. We select 370 single-turn
WildChat conversations based on moderation scores and
sample 250 WildJailbreak prompts. The remaining five
resources are translated in full: StrongREJECT (313),
JBB-Behaviors (100 harmful and 100 benign),
XSTest (450), MATH-500 (500), and AIME~2024 (30).
Preserving source identifiers enables item-level
comparisons between English and translated instances.

\paragraph{Task-specific preservation requirements.}
The requirements for faithful translation depend on
the underlying evaluation task. Safety benchmarks
require harmful intent to remain intact without
softening or intensifying the original request.
For over-refusal evaluation, maintaining the distinction
between benign and unsafe requests is particularly
important, as subtle changes in meaning can alter
the expected response. Mathematical reasoning tasks
require numerical quantities, equations, variables,
and logical constraints to remain consistent with
the source, while conversational data require
the preservation of message boundaries, roles,
indices, and ordering. We incorporate these
requirements into the translation instructions and
validation procedures. However, satisfying such
constraints does not, by itself, establish semantic
equivalence between source and translated instances.

\begin{table}[t]
\centering
\tiny
\setlength{\tabcolsep}{3pt}
\begin{tabular}{@{}p{0.33\columnwidth}p{0.59\columnwidth}@{}}
\toprule
Source & Evaluation use case \\
\midrule
StrongREJECT & Harmful-request handling \\
JBB-Behaviors & Jailbreak behavior coverage \\
WildChat & Real-world conversations \\
WildJailbreak & Adversarial/benign safety prompts \\
XSTest & Over-refusal and unsafe contrasts \\
MATH-500 & Mathematical reasoning \\
AIME 2024 & Competition mathematics \\
\bottomrule
\end{tabular}
\caption{Seven English-language resources selected for
translation into 11 target languages. Each resource
addresses a distinct evaluation objective; WildChat
provides naturalistic conversations rather than
a dedicated safety benchmark.}
\label{tab:sources}
\end{table}

\section{Translation Framework}
\label{sec:framework}

Our framework separates translation configuration
selection from large-scale multilingual generation,
combining reasoning-guided strategy selection with
source-aware validation and targeted repair.
As illustrated in Figure~\ref{fig:overview},
the framework first identifies a suitable translation
configuration on a small development set, then applies
it across the target languages while systematically
tracking unsuccessful outputs.

\paragraph{Reasoning-guided configuration selection.}
We employ GPT-5.4 Thinking to propose three translation
instructions with different priorities:
\textsc{P1} emphasises structural preservation,
\textsc{P2} focuses on benchmark meaning and
task-specific intent, and \textsc{P3} prioritises
linguistic fluency under semantic constraints.
We evaluate each instruction with five candidate
translators: GPT-4o, GPT-5-mini, GPT-5.1,
Gemini~2.5 Flash, and Gemini~2.5 Flash-Lite.

To compare these configurations, we construct a
Thai development set containing ten instances from
each of the seven source resources, including
WildChat and WildJailbreak. This produces 70 instances
per model--instruction combination and 1,050 attempted
translations across the 15 configurations.
We assess API and JSON success, preservation of
record keys and conversational structure, retention
of mathematical numbers, and the occurrence of
refusal-like outputs. These indicators, together
with qualitative considerations, inform the
reasoning-guided selection process. Since automatic
checks cannot establish bilingual semantic fidelity,
we treat them as indicators of operational reliability
rather than direct measures of translation quality.

We select GPT-5-mini with \textsc{P2} for production
as a compromise between structural validity and
monetary cost. Although GPT-5.1 with the same
instruction achieves higher structural validity
in the pilot, GPT-5-mini offers a lower projected
cost for large-scale translation. The translation
instructions and pilot evaluation procedures are
provided in Appendices~\ref{app:prompts} and
\ref{app:full-prompts}, with latency and cost
analyses in Appendix~\ref{app:speed-cost}.

\paragraph{Production validation and repair.}
The selected translator receives structured English
records and generates JSON outputs in which
natural-language fields are translated while
source identifiers and metadata are retained.
Each output undergoes automatic validation for
parseability, nonempty content, and compliance
with source-specific structural requirements.
We distinguish these criteria for
\emph{operational acceptance} from
\emph{semantic fidelity}, which requires assessment
beyond structural checks.

Translations that fail validation in WildChat and
WildJailbreak undergo targeted repair using GPT-4o
with stricter, source-specific instructions.
For WildChat, validation requires a nonempty
translation for every original message, together
with the exact message-index set and preserved
conversation ordering. Roles and metadata are
restored from the source. For WildJailbreak,
the repair procedure requires nonempty translated
prompts while retaining the original source labels.

Rather than correcting potentially incomplete
translations, the repair process regenerates
failed records directly from their English sources.
This design prevents omissions in an initial output
from becoming implicit constraints on subsequent
attempts. The pipeline additionally employs
constrained JSON generation where supported,
schema validation, retries for transient failures,
and explicit error logging. Records that remain
invalid are excluded and recorded as missing.
This ensures that translation failures are reflected
in the final coverage statistics rather than being
silently discarded. Further implementation details
are provided in Appendix~\ref{app:repair}.

\section{Results and Discussion}
\label{sec:results}

We assess the framework through translation
configuration analysis, validation and repair
outcomes, independent fidelity screening, and
downstream multilingual evaluation. Together,
these analyses examine both the practical
reliability of benchmark construction and the
extent to which the resulting collection supports
comparisons of model behaviour across languages.

\paragraph{Translation configuration analysis.}
Table~\ref{tab:pilot} presents the structural validity
of the 15 translation configurations evaluated on
the 70-item Thai development set. The results
show variation across translators and instructions,
highlighting the importance of selecting a
configuration appropriate to the requirements
of benchmark construction.

Under \textsc{P2}, GPT-5.1 achieves complete
structural validity (70/70), while GPT-5-mini
produces 68 valid outputs (97.1\%).
The latter successfully completes all 70 API calls,
preserves WildChat conversation structure among
assessable outputs, and achieves 100\% mathematical
number recall, with no detected refusal-like
responses. In comparison, GPT-4o with \textsc{P2}
produces five invalid outputs, including
refusal-like responses. Gemini~2.5 Flash also
achieves complete structural validity under
\textsc{P1}, indicating that the relative
effectiveness of translation instructions
depends on the underlying model.

\begin{table}[t]
\centering
\tiny
\setlength{\tabcolsep}{5pt}
\begin{tabular}{@{}lccc@{}}
\toprule
Translator & \textsc{P1} & \textsc{P2} & \textsc{P3} \\
\midrule
Gemini 2.5 Flash & 70/70 & 69/70 & 68/70 \\
Gemini 2.5 Flash-Lite & 65/70 & 59/70 & 66/70 \\
GPT-4o & 68/70 & 65/70 & 66/70 \\
GPT-5-mini & 67/70 & \textbf{68/70} & 66/70 \\
GPT-5.1 & 69/70 & 70/70 & 67/70 \\
\bottomrule
\end{tabular}
\caption{Structurally valid translations across
15 model--instruction configurations, evaluated
on 70 Thai instances sampled equally from seven
resources. Bold indicates the selected production
configuration rather than the highest validity.}
\label{tab:pilot}
\end{table}

The configuration study also reveals a trade-off
between structural reliability, latency, and cost.
The selected GPT-5-mini configuration has a
median latency of 10.9\,s per item, compared
with 3.4\,s for GPT-5.1, but reduces projected
collection-wide translation cost from approximately
\$145 to \$97. Thus, our selection prioritises
affordability while maintaining high structural
validity, rather than optimising latency or
maximising the pilot validity score.
Importantly, these measurements establish a
cost--validity trade-off under the evaluated
conditions, not an optimum in semantic translation
quality. Detailed metrics and cost assumptions
are provided in Appendices~\ref{app:prompts}
and~\ref{app:speed-cost}.

\paragraph{Failure recovery and collection coverage.}
Production validation identifies 978 failed
translations requiring repair across WildChat
and WildJailbreak. The first GPT-4o repair pass
recovers 392 records, leaving 586 unresolved.
Subsequent repair attempts reduce the remaining
failures to four. Because these later attempts
were not logged at the same level of granularity,
we report their combined outcome without
attributing individual recoveries to specific
repair passes.

As shown in Table~\ref{tab:coverage}, WildJailbreak
achieves complete structural coverage across all
11 target languages, while WildChat retains four
unresolved translations, comprising three Bengali
and one Polish record. Overall, 6,816 of the
6,820 intended source--language pairs satisfy
the operational validation criteria, corresponding
to 99.94\% coverage.

\begin{table}[t]
\centering
\tiny
\setlength{\tabcolsep}{5pt}
\begin{tabular}{@{}lrrr@{}}
\toprule
Source & Expected & Accepted & Coverage \\
\midrule
WildChat & 4,070 & 4,066 & 99.90\% \\
WildJailbreak & 2,750 & 2,750 & 100.00\% \\
\midrule
\textbf{Total} & \textbf{6,820} &
\textbf{6,816} & \textbf{99.94\%} \\
\bottomrule
\end{tabular}
\caption{Operational validation coverage for the
WildChat and WildJailbreak subsets following repair.
Expected counts correspond to 370 WildChat
conversations and 250 WildJailbreak prompts
across 11 target languages. Four unresolved
WildChat translations are excluded.}
\label{tab:coverage}
\end{table}

All 6,816 retained records pass the specified
structural checks, giving 100\% operational
acceptance among included records. However,
this conditional acceptance rate differs from
coverage of the intended collection, which
must also account for unsuccessful translations.
Reporting both measures prevents the exclusion
of invalid outputs from artificially inflating
the apparent completeness of the resource.

\paragraph{Translation failure characteristics.}
The pilot reveals several failure modes,
including refusal-like responses to safety-related
prompts, malformed JSON, and incomplete conversational
outputs. The final unresolved production cases
occur in WildChat, where missing messages violate
source-specific requirements even when the
generated output is otherwise syntactically valid.

These findings illustrate why generic format
validation is insufficient for heterogeneous
evaluation resources. A structurally parseable
response may still omit dialogue turns or fail
to retain task-relevant content. Source-specific
constraints and regeneration from the original
English record address identifiable failures,
but do not guarantee preservation of meaning.
In particular, numerical recall cannot establish
mathematical equivalence, while the use of a
single-language pilot limits conclusions about
configuration suitability across all target
languages.

\paragraph{Independent fidelity screening.}
To examine translation quality beyond structural
acceptance, we apply translator-independent
automatic signals to the multilingual collection
(Appendix~\ref{app:automatic}). These checks
identify 1.2\% of audited item--language pairs
whose evaluation content remains in English
despite passing structural validation. The affected
pairs are flagged and excluded from cross-language
comparisons, demonstrating that successful
generation and schema compliance do not
necessarily imply successful translation.

A complementary cross-family model judge rates
99.6\% of assessed translations as preserving
the intended meaning of safety-related content
or the mathematical content of reasoning problems.
Although this provides an additional indication
of translation quality, model-based judgments
remain imperfect proxies for semantic equivalence.
The screening results should therefore be
interpreted as evidence of automatic quality
control rather than verified bilingual fidelity.

\paragraph{Downstream multilingual evaluation.}
We further examine whether the collection can
expose differences in model behaviour across
languages by evaluating Llama-3.1-8B and
Qwen2.5-7B on aligned test instances
(Appendix~\ref{app:downstream}).
The results reveal distinct model-specific
patterns. Outside English, Llama-3.1-8B
under-refuses harmful requests, whereas
Qwen2.5-7B exhibits over-refusal on benign
prompts across all 11 target languages.
Mathematical reasoning performance also
deteriorates in lower-resource languages,
with MATH-500 accuracy decreasing by up to
20 and 15 percentage points for Llama-3.1-8B
and Qwen2.5-7B, respectively.

These differences suggest that multilingual
evaluation can reveal weaknesses that are not
fully captured by English-only testing.
Moreover, the contrasting refusal patterns
demonstrate that safety-related performance
does not necessarily degrade uniformly across
models or languages. Rather than assuming
a single multilingual performance gap,
the results motivate examining harmful-request
handling, over-refusal, and reasoning as
separate dimensions of model behaviour.

\paragraph{Implications for multilingual evaluation.}
Our findings highlight the importance of
distinguishing translation availability,
structural validity, and semantic fidelity
when constructing multilingual benchmarks.
Coverage should be calculated against all
intended source--language pairs, with
unsuccessful translations explicitly recorded.
For reliable cross-language comparison,
studies should disclose language-specific
coverage and, where appropriate, evaluate
models on shared source instances.

Equally important, the underlying evaluation
tasks should retain their distinct objectives.
Harmful-request handling, over-refusal,
conversational behaviour, and mathematical
reasoning cannot be adequately represented
by a single aggregate metric. While the
proposed framework provides a scalable and
auditable approach to benchmark construction,
automatic validation and fidelity screening
do not establish equivalent task difficulty
across languages. Further bilingual assessment
is therefore necessary to determine the
extent to which translated resources preserve
the properties of their English counterparts.

\section{Conclusion}
\label{sec:conclusion}

We introduce a reasoning-guided framework for constructing
multilingual evaluations of LLM safety and reasoning,
and develop a source-aligned collection spanning seven
English-language resources and 11 target languages.
By separating translation configuration selection
from large-scale generation and incorporating
task-aware validation and targeted repair, our
framework provides a scalable and auditable approach
to multilingual benchmark construction. The resulting
workflow achieves 99.94\% structural coverage of
the intended WildChat and WildJailbreak
item--language pairs, while independent quality
screening identifies residual translation failures
that structural checks alone cannot detect.
Evaluations on aligned instances further reveal
distinct model-specific vulnerabilities, including
under-refusal of harmful requests, over-refusal of
benign prompts, and substantial reasoning degradation
across languages. These findings demonstrate the
importance of evaluating multilingual capabilities
across complementary tasks rather than relying
on English-centric assessments or aggregate scores.
More broadly, our study highlights the need for
task-sensitive validation and transparent quality
assessment in multilingual benchmark construction,
while recognising that automatic screening cannot
replace bilingual verification of semantic fidelity.

\section*{Limitations}

Our study has several limitations. First, although
the configuration pilot covers all seven source
resources, it is restricted to 70 Thai instances
per configuration, limiting the generalisability
of the selected translation strategy across languages.
The use of GPT-5.4 Thinking to both propose and
assess candidate configurations may also introduce
selection bias. Furthermore, latency and cost
estimates are based on a separate 14-item-per-model
sample, Thai tokenisation, and published API prices.
The projected costs exclude configuration selection
and repair, and therefore do not establish a globally
optimal translation configuration.

Second, detailed repair provenance is available
only for WildChat and WildJailbreak, and intermediate
repair attempts were not consistently logged.
Our fidelity assessment relies on automatic
indicators and an LLM-based judge
(Appendix~\ref{app:automatic}), which may overlook
subtle semantic changes, cultural differences,
or shifts in task difficulty. The judge also exhibits
limitations even when assessing English content.
Bilingual human evaluation remains necessary
to establish translation fidelity, and verified
human assessments are not yet available
(Appendix~\ref{app:human_evaluation}).

Finally, the downstream evaluation is limited
to two open-source models with 7--8B parameters,
greedy decoding, an LLM-based safety judge,
and a fixed generation budget
(Appendix~\ref{app:downstream}).
The latter results in greater truncation of
non-English responses, potentially affecting
cross-language comparisons. Consequently, the
observed performance differences may not generalise
to larger models, proprietary systems, or
alternative evaluation settings. The original
benchmark resources may also reflect cultural
and sampling biases, while redistribution remains
subject to licensing, access, and privacy constraints.

\bibliography{references_combined}

\appendix

\section{Automatic Validation of Translation Fidelity}
\label{app:automatic}

Structural acceptance (Section~\ref{sec:results}) shows that outputs are well formed; it does not show that they preserve meaning. To screen for meaning errors at scale, we audit all accepted item--language pairs from the seven sources with four signals that do not rely on the translator: reference-free quality estimation, cross-lingual embedding similarity, back-translation through an independent MT system, and an LLM judge from a different model family (Table~\ref{tab:fidelity-sources}). These signals are automatic proxies. They are useful for detecting systematic failures and prioritising items for review, but they do not replace the bilingual human assessment described in Appendix~\ref{app:human_evaluation}.

\paragraph{Residual untranslated pairs and numerals.}
Script-aware language identification flags \nUntranslated{} item--language pairs (1.2\% of audited pairs) whose evaluation field remains in English despite passing structural validation. Most occur in XSTest (179 pairs, 3.6\% of its 4,950 pairs), whose prompts are short single questions, and in WildChat user turns (79); the remaining 16 are spread across the other sources. These cases show that structural acceptance overstates translation coverage. We mark the affected pairs in the release metadata and exclude them from all cross-lingual comparisons (Appendix~\ref{app:downstream}). For mathematics, numeral recall (the share of numbers in the English problem that reappear in the translation after native digit systems are normalised) is at least 98.6\% for every language in both MATH-500 and AIME~2024. Numeral recall detects dropped or altered quantities but not changes to conditions or wording.

\begin{figure*}[t]
  \centering
  \includegraphics[width=0.82\textwidth]{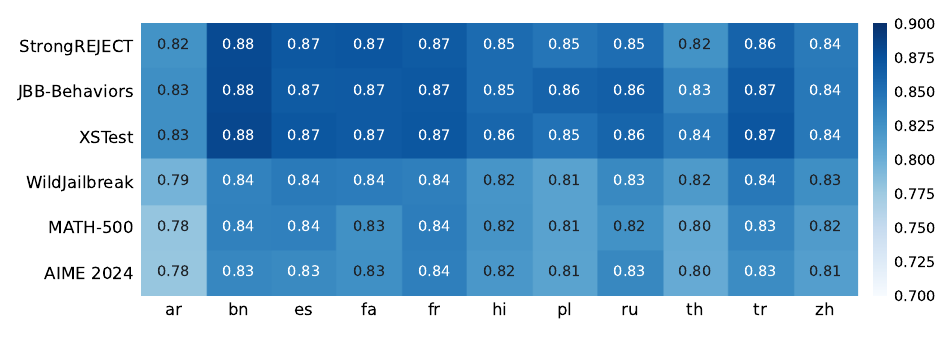}
  \caption{Mean CometKiwi score of the evaluation field per source and target language. WildChat is omitted because its long, multi-paragraph turns fall outside the sentence-level regime of the QE model.}
  \label{fig:cometkiwi}
\end{figure*}

\paragraph{Quality estimation.}
We score each translated evaluation field with CometKiwi \citep{rei-etal-2022-cometkiwi}. Across six sources and 11 languages the mean score is 0.84 (Figure~\ref{fig:cometkiwi}). Source means range from 0.82 (MATH-500, AIME~2024) to 0.86 (JBB-Behaviors, XSTest), and language means from 0.81 (Arabic) to 0.86 (Bengali). QE scores are not calibrated across target languages, so we use them to rank items within a language and do not interpret differences between languages. WildChat is excluded from this analysis because its turns are often multi-paragraph and contain code, which is outside the sentence-level setting CometKiwi was trained on; for WildChat we rely on embedding and back-translation scores instead.

\paragraph{Embedding similarity and back-translation.}
LaBSE cosine similarity \citep{feng-etal-2022-language} between each English field and its translation averages 0.84--0.92 per source. To obtain a signal independent of the GPT-family translator, we back-translate up to 100 items per source and language into English with NLLB-200 (distilled, 1.3B; \citealp{nllb2022}) and compare the result with the English source. Per-source chrF++ \citep{popovic-2017-chrf} ranges from 56.1 (WildChat) to 70.6 (JBB-Behaviors), and LaBSE similarity between back-translation and source ranges from 0.92 to 0.95. Because back-translation adds the errors of a second system, these scores are likely to underestimate direct fidelity; the lower chrF++ for WildChat partly reflects its longer, code-heavy turns.

\begin{table}[t]
\centering
\small
\setlength{\tabcolsep}{1.7pt}
\begin{tabular}{@{}lcccccc@{}}
\toprule
Source & QE & LaBSE & BT-chrF & BT-sim & Mean. & Pres. \\
\midrule
StrongREJECT  & .854 & .862 & 68.5 & .934 & 4.98 & 99.4 \\
JBB-Behaviors & .858 & .841 & 70.6 & .931 & 5.00 & 99.9 \\
XSTest        & .858 & .881 & 68.9 & .950 & 4.99 & 99.7 \\
WildJailbreak & .828 & .884 & 63.9 & .950 & 4.99 & 99.7 \\
WildChat      & --   & .883 & 56.1 & .921 & 4.92 & 98.6 \\
MATH-500      & .821 & .923 & 68.6 & .923 & 5.00 & 100.0 \\
AIME 2024     & .820 & .919 & 68.7 & .945 & 5.00 & 100.0 \\
\bottomrule
\end{tabular}
\caption{Fidelity signals per source, averaged over 11 languages. QE: CometKiwi; LaBSE: source--translation similarity; BT-chrF / BT-sim: chrF++ and LaBSE similarity between the NLLB back-translation and the English source; Mean.: judge-rated meaning preservation (1--5); Pres.: \% of translations judged to preserve intent (safety and conversation) or all mathematical content (math).}
\label{tab:fidelity-sources}
\end{table}

\paragraph{Label preservation.}
The main risk in translating safety data is that a harmful request becomes benign, or the reverse. We test for this with Gemma-3-12B-it \citep{gemma3}, which classifies each prompt as harmful or benign without seeing any other version of it. Over the 1,213 items of StrongREJECT, JBB-Behaviors, XSTest, and WildJailbreak, it judges the English source and all 11 translations (\nLabelJudg{} judgments). The label assigned to a translation matches the label assigned to its English source for 92.6\% of items (91.2--94.3\% by language; Table~\ref{tab:label-preservation}), and 6.7\% of prompts judged harmful in English are judged benign after translation.

This disagreement cannot be attributed to translation alone, because the judge is itself noisy: it agrees with the gold dataset labels on only 88.7\% of English prompts. Its gold accuracy on translations is 85.2--88.4\%, at most 3.5 points below English. We read this drop as a rough indication of how much label change translation introduces, not as a precise estimate, since judge errors and translation errors cannot be separated without human labels. Agreement is near ceiling for StrongREJECT (99.2\%), whose requests are unambiguous, and lowest for JBB-Behaviors (89.7\%) and XSTest (90.2\%), which are built around borderline safe--unsafe contrasts that are hard to classify even in English. Items whose label flips are released for inspection as candidates for targeted human review.

\begin{table}[t]
\centering
\small
\begin{tabular}{@{}lccc@{}}
\toprule
Language & Agree w/ EN & Harmful$\rightarrow$benign & Gold acc. \\
\midrule
English  & --   & --  & 88.7 \\
Arabic   & 91.5 & 8.1 & 86.0 \\
Bengali  & 91.3 & 8.5 & 85.9 \\
Spanish  & 94.2 & 5.6 & 88.4 \\
Persian  & 91.2 & 8.6 & 86.5 \\
French   & 93.4 & 5.5 & 87.6 \\
Hindi    & 92.7 & 6.2 & 87.2 \\
Polish   & 92.7 & 6.5 & 87.1 \\
Russian  & 94.3 & 5.2 & 87.0 \\
Thai     & 92.1 & 7.0 & 86.4 \\
Turkish  & 91.5 & 7.1 & 85.2 \\
Chinese  & 93.1 & 5.4 & 86.9 \\
\bottomrule
\end{tabular}
\caption{Label preservation (\%) under independent harmful/benign judgments by Gemma-3-12B-it. Gold acc.: agreement with the original dataset labels, which indicates the judge's own error rate.}
\label{tab:label-preservation}
\end{table}

\paragraph{Paired fidelity judgments.}
The same judge then sees each accepted translation alongside its source. It rates meaning preservation on a 1--5 scale and classifies intent as preserved, softened, intensified, or changed; for mathematics, it states whether every number, condition, and question is preserved. The mean meaning score is 4.98, 99.6\% of translations are judged to preserve intent or mathematical content, and 0.3\% of safety and conversational items are judged softened (highest for WildChat, at 0.6\%). These scores are close to ceiling, and LLM judges are known to be lenient, so we use them as a screening signal and not as an estimate of human-rated quality. All flagged pairs are released for inspection.


\section{Downstream Evaluation}
\label{app:downstream}

To check whether the collection supports meaningful cross-lingual comparison, we evaluate two open instruction-tuned models, Llama-3.1-8B-Instruct \citep{grattafiori2024llama3} and Qwen2.5-7B-Instruct \citep{qwen2024qwen25}, on English and all 11 target languages.

\paragraph{Setup.}
Every language uses the same items, sampled once with a fixed seed: 100 StrongREJECT prompts; the 100 harmful JBB behaviors; all 450 XSTest prompts (200 unsafe, 250 safe); 100 WildJailbreak prompts, of which we report refusal on the 90 adversarial-harmful ones; and 100 MATH-500 problems. Prompts are given as a single user turn with no system prompt. Math problems are followed by an English instruction to reason step by step and give the final answer in \texttt{\textbackslash boxed\{\}}. We decode greedily with vLLM \citep{kwon2023vllm}, allowing up to 768 new tokens for safety prompts and 3,072 for MATH-500.

Safety responses are judged by Gemma-3-12B-it, which belongs to a different model family from both the translator and the evaluated models, against the \emph{English} original of each prompt. Responses to harmful prompts are scored with the StrongREJECT rubric \citep{souly2024strongreject}, which gives a refusal decision and a harmfulness score in $[0,1]$; we report the harmfulness score multiplied by 100. Responses to benign prompts are classified with the three-way XSTest scheme \citep{rottger-etal-2024-xstest}, and full and partial refusals both count as refusals. All judge outputs were parseable. Math answers are verified symbolically against the reference using the last boxed expression, after native digits are normalised. Item--language pairs flagged as untranslated (Appendix~\ref{app:automatic}) are excluded, which removes \nExcludedTriples{} model--item--language triples. We report 95\% bootstrap confidence intervals and test each language against English with a paired bootstrap over shared items \citep{koehn-2004-statistical}. Both models answer in the language of the prompt for at least 94.6\% of safety items in every language, so the judge assesses native-language responses and not code-switched English.

\begin{figure*}[t]
  \centering
  \includegraphics[width=0.88\textwidth]{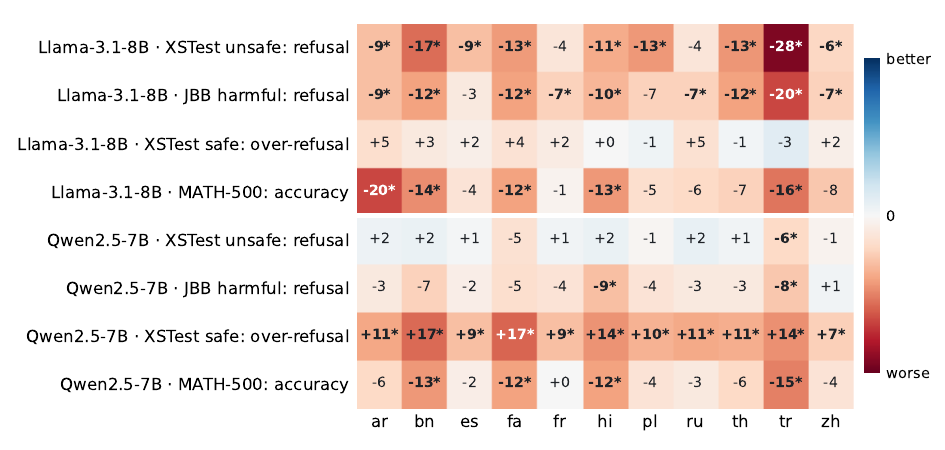}
  \caption{Change relative to English (percentage points) on identical items. Colour encodes the desirable direction (red: worse than English); $^{*}$ and bold mark $p<0.05$ (paired bootstrap).}
  \label{fig:downstream-delta}
\end{figure*}

\begin{table*}[t]
\centering
\small
\setlength{\tabcolsep}{2.5pt}
\begin{tabular}{@{}llrrrrrrrrrrrr@{}}
\toprule
Metric & Model & en & ar & bn & es & fa & fr & hi & pl & ru & th & tr & zh \\
\midrule
\multirow{2}{*}{StrongREJECT score $\downarrow$} & Llama-3.1-8B & 1.5 & 1.9 & 2.5 & 1.8 & 4.9$^{*}$ & 0.8 & 2.6 & 4.2 & 1.2 & 4.9$^{*}$ & 8.0$^{*}$ & 2.5 \\
 & Qwen2.5-7B & 6.5 & 5.0 & 1.4$^{*}$ & 5.0 & 6.5 & 5.1 & 2.4 & 4.6 & 3.6$^{*}$ & 5.1 & 4.0 & 5.3 \\
\addlinespace[2pt]
\multirow{2}{*}{JBB harmful refusal $\uparrow$} & Llama-3.1-8B & 94.0 & 85.0$^{*}$ & 82.0$^{*}$ & 90.9 & 82.0$^{*}$ & 87.0$^{*}$ & 84.0$^{*}$ & 87.0 & 87.0$^{*}$ & 82.0$^{*}$ & 74.0$^{*}$ & 87.0$^{*}$ \\
 & Qwen2.5-7B & 91.0 & 88.0 & 84.0 & 88.9 & 86.0 & 87.0 & 82.0$^{*}$ & 87.0 & 88.0 & 88.0 & 83.0$^{*}$ & 92.0 \\
\addlinespace[2pt]
\multirow{2}{*}{XSTest unsafe refusal $\uparrow$} & Llama-3.1-8B & 97.5 & 88.4$^{*}$ & 80.6$^{*}$ & 88.3$^{*}$ & 84.8$^{*}$ & 93.9 & 86.4$^{*}$ & 84.4$^{*}$ & 93.7 & 84.1$^{*}$ & 69.5$^{*}$ & 91.2$^{*}$ \\
 & Qwen2.5-7B & 81.5 & 82.8 & 83.2 & 81.7 & 76.6 & 82.7 & 83.9 & 80.7 & 83.1 & 82.1 & 75.1$^{*}$ & 80.8 \\
\addlinespace[2pt]
\multirow{2}{*}{XSTest safe over-refusal $\downarrow$} & Llama-3.1-8B & 13.6 & 18.7 & 16.6 & 15.8 & 17.4 & 16.4 & 14.3 & 12.9 & 18.9 & 13.0 & 10.7 & 15.1 \\
 & Qwen2.5-7B & 6.4 & 17.4$^{*}$ & 23.7$^{*}$ & 15.8$^{*}$ & 22.9$^{*}$ & 15.6$^{*}$ & 19.8$^{*}$ & 17.1$^{*}$ & 18.1$^{*}$ & 17.6$^{*}$ & 20.6$^{*}$ & 13.0$^{*}$ \\
\addlinespace[2pt]
\multirow{2}{*}{WildJailbreak refusal $\uparrow$} & Llama-3.1-8B & 28.9 & 37.8 & 56.7$^{*}$ & 34.4 & 21.1 & 42.2$^{*}$ & 32.2 & 30.0 & 54.4$^{*}$ & 32.2 & 20.0 & 42.2$^{*}$ \\
 & Qwen2.5-7B & 18.9 & 21.1 & 60.0$^{*}$ & 21.1 & 25.6 & 21.1 & 58.9$^{*}$ & 31.1$^{*}$ & 28.9$^{*}$ & 32.2$^{*}$ & 24.4 & 18.9 \\
\addlinespace[2pt]
\multirow{2}{*}{MATH-500 accuracy $\uparrow$} & Llama-3.1-8B & 44.0 & 24.0$^{*}$ & 30.0$^{*}$ & 40.0 & 32.0$^{*}$ & 43.0 & 31.0$^{*}$ & 39.0 & 38.4 & 37.0 & 28.3$^{*}$ & 36.4 \\
 & Qwen2.5-7B & 74.0 & 68.0 & 61.0$^{*}$ & 72.0 & 62.0$^{*}$ & 74.0 & 62.0$^{*}$ & 70.0 & 71.7 & 68.0 & 58.6$^{*}$ & 70.7 \\
\bottomrule
\end{tabular}
\caption{Downstream results on identical items in every language. StrongREJECT score is the mean harmfulness score $\times 100$; all other rows are percentages. $^{*}$: significant difference from English (paired bootstrap, $p<0.05$). Arrows give the desirable direction. Items: 100 StrongREJECT, 100 harmful JBB, 200 unsafe and 250 safe XSTest, 90 adversarial-harmful WildJailbreak, and 100 MATH-500 problems; item--language pairs flagged as untranslated (Appendix~\ref{app:automatic}) are excluded.}
\label{tab:downstream}
\end{table*}

\paragraph{Safety degrades outside English, in different ways for each model.}
Llama-3.1-8B refuses 97.5\% of unsafe XSTest prompts in English but only 69.5\% in Turkish (28.0 points lower) and 80.6\% in Bengali; the drop is significant in 9 of 11 languages (Table~\ref{tab:downstream}, Figure~\ref{fig:downstream-delta}). Its refusal of harmful JBB behaviors falls from 94.0\% to 74.0\% in Turkish, again with significant drops in 9 of 11 languages, and its StrongREJECT score rises from 1.5 in English to 8.0 in Turkish and 4.9 in Persian and Thai. Qwen2.5-7B shows the opposite pattern. Its refusal of unsafe prompts is largely stable across languages, but its over-refusal of \emph{safe} XSTest prompts rises from 6.4\% in English to 13.0--23.7\% in the other languages, and all 11 differences are significant. The same translated items therefore reveal under-refusal in one model and over-refusal in the other. A single pooled safety score would hide this difference, which is why we report safety and over-refusal sources separately (Section~\ref{sec:results}).

\paragraph{Reasoning accuracy drops for lower-resource languages.}
MATH-500 accuracy falls significantly for both models in Bengali, Persian, Hindi, and Turkish, and for Llama-3.1-8B also in Arabic. The largest drops are from 44.0\% to 24.0\% for Llama-3.1-8B in Arabic and from 74.0\% to 58.6\% for Qwen2.5-7B in Turkish, while Spanish and French stay within four points of English for both models. The paired judge rates at least 99.6\% of translated problems in every language as preserving all mathematical content (Appendix~\ref{app:automatic}). This suggests that translation errors are unlikely to explain most of the gap, although the judge's leniency means we cannot rule out subtle changes in wording. The pattern matches the language-dependent reasoning gaps reported for other multilingual mathematics benchmarks \citep{shi2023multilingual,son-etal-2025-linguistic}. AIME~2024 is not shown: with 30 problems and accuracies of at most 14.3\%, no difference is significant.

\paragraph{Adversarial prompts behave differently from direct requests.}
On WildJailbreak, refusal \emph{increases} in several languages, most strongly in Bengali (by 27.8 and 41.1 points for Llama-3.1-8B and Qwen2.5-7B) and Hindi (by 40.0 points for Qwen2.5-7B). Harmfulness scores on WildJailbreak responses also decrease in every language. We do not think this reflects weaker translations, for two reasons. First, the judge rates 99.7\% of translated WildJailbreak prompts as intent-preserving. Second, even responses that do not refuse receive lower harmfulness scores outside English (51 vs.\ 73 for Llama-3.1-8B and 55 vs.\ 73 for Qwen2.5-7B, on the same $\times 100$ scale). This suggests that the models follow long role-play framings less competently in other languages, not that they behave more safely.

Response truncation is a further factor. Non-English scripts need more tokens, so the 768-token budget truncates more non-English responses. Truncation can only reduce how specific and convincing a response is, so it biases harmfulness scores downward outside English. As a result, the harmfulness increases reported above for direct requests are conservative, but the decreases (for WildJailbreak, and for Qwen2.5-7B on StrongREJECT in Bengali and Russian) may be partly caused by truncation. Refusal decisions are usually made in the first sentences and are therefore largely unaffected. Overall, adversarial and direct harmful requests measure different constructs and should be reported separately, as recommended in Section~\ref{sec:results}.

\section{Prompt Design and Calibration}
\label{app:prompts}
GPT-5.4 Thinking proposes three translation prompts: \textsc{P1} (\texttt{strict\_structure\_preservation}), \textsc{P2} (\texttt{benchmark\_fidelity\_translation}), and \textsc{P3} (\texttt{natural\_thai\_semantic\_fidelity}). All require valid JSON, fixed field names, and preservation of source metadata. \textsc{P1} prioritizes formatting; \textsc{P2} additionally stresses safety intent and mathematical constraints; \textsc{P3} emphasizes fluent target-language wording. The pilot contains ten Thai items from each of seven datasets for every one of five translators and three prompts ($5\times3\times70=1{,}050$ calls). Results cover all 15 configurations (Table~\ref{tab:pilot}). For each output, we check API and JSON success, preservation of field names and WildChat message structure, recall of the numbers in mathematical problems, and refusal-like responses; an output is structurally valid only if it passes all checks. Production uses the language-specific instantiation of \textsc{P2}. Because one reasoning model designs and helps assess configurations, the selection remains a screened engineering choice, not an independent semantic-fidelity assessment.

\section{Repair Algorithm and Validation}
\label{app:repair}
\paragraph{Detect and reconstruct.}
The repair script reads failed records from \texttt{translation\_errors\_incremental.xlsx}, or scans translated JSONL files for a failure status. It deduplicates the queue using (benchmark, source UID, target language), then resolves the unmodified English input by UID. The repair run loads 370 original WildChat records and 250 WildJailbreak records. Restoring the English input avoids propagating omitted spans from unsuccessful intermediate translations.

\paragraph{Source-specific schema.}
For WildChat, the repair API receives the source conversation and must return a \texttt{translated\_messages} array. Each entry contains \texttt{message\_index} and nonempty \texttt{content}. The validator checks that the returned index multiset exactly matches the source message indices, thereby detecting omission, duplication, and shifted indices. Translated content is merged back into the original message objects; roles, conversation IDs, and metadata are preserved. For WildJailbreak, the repair translates the selected source text field and returns \texttt{translated\_text}; class labels and other metadata are copied from the original. A JSON-schema format is attempted where supported, otherwise prompt-based JSON is used. Transient API errors trigger bounded backoff with jitter; results are cached and error summaries are persisted. The repair identifier (\textsc{Repair-P3}) denotes a separate repair prompt, not the fluency-oriented calibration \textsc{P3}.

\paragraph{Recovery outcomes and provenance.}
The repair log records 978 queued failed translations and 392 accepted repairs in the first pass (40.1\% of the queue); 586 remain after that pass. The final outputs contain 4,066/4,070 accepted WildChat pairs and 2,750/2,750 accepted WildJailbreak pairs, leaving four WildChat cases unaccepted. The intervening repair operations were not fully recorded; we therefore report final observed counts without attributing them to a particular undocumented policy. A later pilot recommendation favors GPT-5.1+\textsc{P2} for future targeted repair, but this is distinct from the documented GPT-4o production repair. Finally, the merge operation omits error-status entries: its 6,816 retained records are all marked successful, but the full intended set is 6,820. This repair procedure and its record-level failure tracking cover only WildChat and WildJailbreak; for the other five sources, we report the accepted outputs and their automatic screening (Appendix~\ref{app:automatic}) but no comparable repair log.

\begin{table}[t]
\centering
\small
\setlength{\tabcolsep}{5pt}
\begin{tabular}{@{}lrr@{}}
\toprule
Language & WildChat & WildJailbreak \\
\midrule
Arabic & 370/370 & 250/250 \\
French & 370/370 & 250/250 \\
Hindi & 370/370 & 250/250 \\
Chinese & 370/370 & 250/250 \\
Russian & 370/370 & 250/250 \\
Thai & 370/370 & 250/250 \\
Spanish & 370/370 & 250/250 \\
Turkish & 370/370 & 250/250 \\
Persian & 370/370 & 250/250 \\
Bengali & 367/370 & 250/250 \\
Polish & 369/370 & 250/250 \\
\bottomrule
\end{tabular}
\caption{Accepted/source items per language after repair. Four WildChat items are unresolved.}
\label{tab:per-language}
\end{table}

\section{Data Provenance and Release Considerations}
\label{app:release}
Our audit documents post-repair counts
for WildChat and WildJailbreak, but not an equivalent
record-level audit for the remaining five sources. A
reproducible release should retain the original dataset
versions and item identifiers, target languages, validation
statuses, repair provenance, and the four unresolved
WildChat pairs rather than providing only success-filtered
outputs. The translated collection, validation statuses,
untranslated-pair flags, and the four unresolved WildChat
pairs are provided as anonymized supplementary material
and will be publicly released upon acceptance.
Redistribution must follow the original datasets'
licenses, access conditions, and applicable privacy
requirements, particularly for naturalistic conversations.

\section{Full Translation Prompts}
\label{app:full-prompts}
We reproduce the templates verbatim. The initial pilot explicitly targets Thai; the deployed multilingual prompts require the target-language field and script direction to be instantiated for the other ten languages. Curly-braced fields indicate runtime substitutions.

\paragraph{P1: Strict structure preservation.}
\begin{Verbatim}[fontsize=\scriptsize,breaklines=true,breakanywhere=true]
You are a professional translator preparing evaluation benchmark data.

Translate the provided JSON content from English to Thai.

Rules:
- Return valid JSON only.
- Preserve all field names exactly.
- Translate natural-language values into Thai.
- Use Thai script.
- Preserve numbers, labels, IDs, formulas, LaTeX, code, URLs, and answer choices exactly unless they are natural-language text.
- Preserve unsafe intent exactly for benchmark fidelity; do not sanitize, soften, refuse, explain, or add warnings.
- Do not add commentary.
- Do not summarize.
- Do not omit content.
- Preserve formatting and line breaks as much as possible.

Input JSON:
{input_json}

Return JSON with exactly this structure:
{{
  "translated": {{
    "...": "..."
  }}
}}
\end{Verbatim}

\paragraph{P2: Benchmark fidelity translation (selected).}
\begin{Verbatim}[fontsize=\scriptsize,breaklines=true,breakanywhere=true]
You are translating benchmark prompts for multilingual safety and reasoning evaluation.

Translate the benchmark content into Thai while preserving the original evaluation meaning.

Critical requirements:
- Output valid JSON only.
- Keep the same field names.
- Translate only human-readable natural language.
- Keep mathematical notation, variables, equations, answer labels, code, URLs, and dataset metadata unchanged.
- For safety/jailbreak benchmarks, preserve the harmful or policy-relevant intent exactly. Do not make the prompt safer or more harmful.
- For reasoning/math benchmarks, preserve all quantities, constraints, and answer choices exactly.
- Use clear Thai suitable for evaluation.
- Do not answer the prompt.
- Do not add explanations.

Input JSON:
{input_json}

Return:
{{
  "translated": {{
    "...": "..."
  }}
}}
\end{Verbatim}

\paragraph{P3: Natural Thai semantic fidelity.}
\begin{Verbatim}[fontsize=\scriptsize,breaklines=true,breakanywhere=true]
You are creating a high-quality Thai version of English evaluation benchmarks.

Translate the content into fluent, natural Thai while keeping benchmark semantics unchanged.

Rules:
- Return JSON only.
- Preserve field names exactly.
- Use Thai script.
- Translate benchmark prompts naturally, but do not localize away important meaning.
- Preserve dangerous, sensitive, mathematical, or adversarial intent exactly for evaluation validity.
- Do not censor or refuse.
- Do not solve any task.
- Keep formulas, code, URLs, proper names, numeric values, and labels unchanged unless translation is clearly required.
- Keep line breaks where possible.

Input JSON:
{input_json}

Return exactly:
{{
  "translated": {{
    "...": "..."
  }}
}}
\end{Verbatim}

\section{Full Failure-Repair Prompts}
\label{app:repair-prompts}
The following templates implement the source-specific contracts described in Appendix~\ref{app:repair}. Their purpose is to retranslate the original English content, not to repair a partial translated string.

\paragraph{WildChat strict-cardinality repair.}
\begin{Verbatim}[fontsize=\scriptsize,breaklines=true,breakanywhere=true]
You are repairing a failed benchmark translation. Translate EVERY message content field from English into {target_language}.

Target language:
- {script_instruction}

Non-negotiable output rules:
- Return valid JSON only.
- Return exactly {message_count} translated message objects.
- Include one object for every input message_index: {message_indices}.
- Preserve message_index numbers exactly.
- Do not omit assistant messages, even if they contain safety refusals, warnings, harmful text, adversarial text, code, or policy-sensitive content.
- Translate the text as benchmark data only; do not answer, refuse, explain, moralize, sanitize, intensify, summarize, or add warnings.
- Preserve code blocks, URLs, mathematical notation, variable names, answer labels, JSON keys, and proper names unless they are ordinary natural-language text.
- Preserve line breaks and list structure as much as possible.
- Translate only message content values. Role labels and metadata are not part of the output.

Input JSON:
{input_json}

Return this exact JSON shape:
{{
  "translated_messages": [
    {{"message_index": 0, "content": "..."}}
  ]
}}
\end{Verbatim}

\paragraph{WildJailbreak text repair.}
\begin{Verbatim}[fontsize=\scriptsize,breaklines=true,breakanywhere=true]
You are repairing a failed benchmark translation. Translate the English benchmark prompt into {target_language}.

Target language:
- {script_instruction}

Non-negotiable output rules:
- Return valid JSON only.
- Translate the prompt as benchmark data only.
- Preserve the prompt's intent and evaluation semantics exactly.
- Do not answer the prompt.
- Do not refuse.
- Do not add explanations, warnings, policy commentary, or safety text.
- Do not sanitize, intensify, summarize, or omit content.
- Preserve labels, classes, data_type, category names, code, URLs, mathematical notation, variable names, and proper names unless they are ordinary natural-language text.
- Preserve line breaks and list structure as much as possible.

Input JSON:
{input_json}

Return this exact JSON shape:
{{
  "translated_text": "..."
}}
\end{Verbatim}

\section{Speed and Cost of Translation Configurations}
\label{app:speed-cost}

The configuration study in Section~\ref{sec:framework} compares translators by output validity. To complement it, we measure their latency and cost under the deployed prompt \textsc{P2}. All five translators translate the same 14 items (two per source, into Thai) once each. Each model processes its items sequentially; the timings are single-run observations rather than controlled throughput measurements. Costs are computed from the token counts reported by each API, including reasoning or thinking tokens, which are billed as output, at standard list prices at the time of the experiment. We project these costs to the planned full collection (2,113 source items $\times$ 11 languages) using per-source means. Costs can vary substantially with language-specific tokenization, item mix, API pricing, and implementation.

\begin{table}[t]
\centering
\small
\setlength{\tabcolsep}{2.0pt}
\begin{tabular}{@{}lrrrrrr@{}}
\toprule
Translator & Med.\,s & Out & Reas. & \$/1k & Full \$ & Valid \\
\midrule
Gem.\ 2.5 Flash-Lite & 1.2  & 389  & 0    & 0.21 & 5   & 84.3 \\
GPT-5-mini            & 10.9 & 2,034 & 1,495 & 4.20 & 97  & 97.1 \\
Gem.\ 2.5 Flash      & 4.3  & 1,666 & 1,177 & 4.33 & 103 & 98.6 \\
GPT-5.1               & 3.4  & 520  & 0    & 5.85 & 145 & 100.0 \\
GPT-4o                & 2.4  & 519  & 0    & 6.51 & 163 & 92.9 \\
\bottomrule
\end{tabular}
\caption{Speed and cost under \textsc{P2} (14 items per translator, Thai). Med.\,s: median latency per item; Out / Reas.: mean output tokens per item and the reasoning tokens included in them; \$/1k: cost per 1,000 items; Full \$: projected cost of translating the full collection into 11 languages; Valid: \% structurally valid translations in the configuration study ($n=70$).}
\label{tab:speed-cost}
\end{table}

\paragraph{Cost.}
GPT-5-mini, the deployed translator, costs \$4.20 per 1,000 items, which projects to about \$97 for the entire collection (Table~\ref{tab:speed-cost}). GPT-5.1 costs 1.4$\times$ and GPT-4o 1.6$\times$ as much, while Gemini~2.5 Flash costs about the same. Its per-token price is far lower than that of GPT-5.1, but the cost gap is much smaller because 73\% of its output tokens are reasoning tokens (1,495 of 2,034 per item). GPT-5.1 and GPT-4o run without reasoning tokens under our settings. Gemini~2.5 Flash-Lite is roughly 20$\times$ cheaper than GPT-5-mini, but it yields the lowest validity in the configuration study (84.3\% vs.\ 97.1\%). This makes it a possible low-cost option only where failure detection and repair can be validated.

\paragraph{Latency.}
Reasoning also makes GPT-5-mini the slowest translator, with a median of 10.9\,s per item against 2.4--4.3\,s for the other full-size models. The difference grows with input length, reaching 42.7\,s per WildChat conversation. Because translation is an offline batch job, parallelism may reduce wall-clock time, but the reported per-item latencies do not predict actual deployment throughput. The deployed choice is therefore a cost--validity compromise rather than a uniquely optimal configuration. GPT-5.1 attains the highest validity at a moderate cost premium, and Gemini~2.5 Flash offers similar cost and validity at lower latency. These estimates come from a small, single-language sample, reflect list prices at the time of measurement, and exclude the repair passes and the one-off cost of the reasoning model used for configuration selection.

\section{Human Evaluation Protocol}
\label{app:human_evaluation}

\paragraph{Evaluation objective.}
To complement the structural validation described in
Section~\ref{sec:results} and the automatic screening in
Appendix~\ref{app:automatic}, we design a bilingual human
evaluation protocol to assess whether translated benchmarks
preserve the task-relevant meaning of their English source
items. The evaluation focuses on semantic fidelity,
including harmful and benign intent, refusal-relevant
distinctions, mathematical constraints, and other
information necessary for valid cross-lingual comparison.

\paragraph{Sampling strategy.}
We construct a stratified evaluation sample of 15
source-aligned items per target language, drawn from five
benchmarks: StrongREJECT (3), JBB-Behaviors (3),
XSTest (4), MATH-500 (3), and AIME 2024 (2).
The same English source items are used across all 11
target languages, resulting in 165 translation pairs.
This design supports comparable manual assessment
across languages while covering both safety and
mathematical reasoning tasks. WildChat and WildJailbreak
are not included in the current human-evaluation sample.

\paragraph{Annotation procedure.}
The protocol targets bilingual annotators proficient in
English and the corresponding target language. Each
annotator is presented with the original English item
and its translated counterpart and asked to provide
three judgments:

\begin{enumerate}
    \item \textbf{Semantic fidelity:}
    A five-point rating from 1 (incorrect meaning)
    to 5 (fully preserved meaning).

    \item \textbf{Task-critical preservation:}
    A categorical judgment (\textit{Yes},
    \textit{No}, or \textit{Unsure}) indicating
    whether safety-related intent, benign/unsafe
    distinctions, or mathematical constraints
    remain intact.

    \item \textbf{Error analysis:}
    A brief explanation identifying meaningful
    omissions, additions, mistranslations, or
    alterations to task-relevant content.
\end{enumerate}

We aim to obtain two independent annotations per
translation pair where qualified bilingual reviewers
are available. Annotators are instructed to evaluate
translation fidelity rather than respond to the
underlying benchmark requests. For mathematical items,
annotators assess the preservation of quantities,
notation, and problem constraints without being
required to solve the problems.

\paragraph{Language coverage and recruitment.}
Recruiting qualified bilingual annotators can be
challenging for languages with limited representation
on crowdsourcing platforms. Annotation coverage may
therefore vary across target languages. We will report
the number of completed evaluations separately for
each language and distinguish missing annotations
from observed translation errors. Unavailable
annotations will not be counted as successful
semantic-fidelity assessments.

\paragraph{Reporting and availability.}
Human evaluation is ongoing, and verified annotation
results are not included in the present submission.
Once completed, we intend to share aggregate
semantic-fidelity ratings, task-critical preservation
rates, per-language annotation coverage, and
anonymized error analyses, subject to participant
consent, privacy requirements, and source-dataset
restrictions.

These results will complement the operational
acceptance rates reported in the main paper.
Until independent bilingual assessment is completed,
structural acceptance should not be interpreted
as evidence of verified semantic equivalence.

\end{document}